\documentclass[review]{elsarticle}       
\usepackage{hhline}
\usepackage{natbib}
\usepackage{palatino,epsfig,latexsym}
\usepackage{multirow}
\usepackage{array}
\usepackage[table]{xcolor}
\usepackage{longtable}
\usepackage{makecell}
\usepackage{enumitem}
\usepackage{pdflscape}
\graphicspath{{./4_cMLSGA_Figures/}}

\definecolor{redbox}{RGB}{244,176,131}
\definecolor{greenbox}{RGB}{168,208,141}
\definecolor{yellowbox}{RGB}{255,217,102}
\definecolor{bluebox}{RGB}{124,175,222}
\usepackage{lineno,hyperref}

\begin{document}

			\begin{frontmatter}
				
				\title{ cMLSGA: A Co-Evolutionary Multi-Level Selection Genetic Algorithm for Multi-Objective Optimization}

				\author[address1]{ P. A. Grudniewski\corref{mycorrespondingauthor}}
				\cortext[mycorrespondingauthor]{Corresponding author}
				\ead{pag1c18@soton.ac.uk}
				\author[address1,address2]{A. J. Sobey}
				\ead{ajs502@soton.ac.uk}
				
				\address[address1]{Fluid Structure Interactions Group, University of Southampton, Southampton, SO16 7QF, England, UK}
				\address[address2]{	Marine and Maritime Group, Data-centric Engineering, The Alan Turing Institute, The British Library, London, NW1 2DB, England, UK}
				
			\begin{abstract}
				In practical optimisation the dominant characteristics of the problem are often not known prior. Therefore, there is a need to develop general solvers as it is not always possible to tailor a specialised approach to each application. The hybrid form of Multi-Level Selection Genetic Algorithm (MLSGA) already shows good performance on range of problems due to its diversity-first approach, which is rare among Evolutionary Algorithms. To increase the generality of its performance this paper proposes a distinct set of co-evolutionary mechanisms, which defines co-evolution as competition between collectives rather than individuals. This distinctive approach to co-evolutionary provides less regular communication between sub-populations and different fitness definitions between individuals and collectives. This encourages the collectives to act more independently creating a unique sub-regional search, leading to the development of co-evolutionary MLSGA (cMLSGA). To test this methodology nine genetic algorithms are selected to generate several variants of cMLSGA, which incorporates these approaches at the individual level. The new mechanisms are tested on over 100 different functions and benchmarked against the 9 state-of-the-art competitors in order to find the best general solver. The results show that the diversity of co-evolutionary approaches is more important than their individual performances. This allows the selection of two competing algorithms that improve the generality of cMLSGA, without large loss of performance on any specific problem type. When compared to the state-of-the-art, the proposed methodology is the most universal and robust, leading to an algorithm more likely to solve complex problems with limited knowledge about the search space. 
				\end{abstract}
				
				\begin{keyword}
				Genetic algorithms\sep co-evolutionary\sep multi-level selection\sep multi-objective optimisation.
				\end{keyword}
				
			\end{frontmatter}

\section{The requirement for general algorithms}

There is a growing interest in computer-aided optimisation of practical problems. This has driven the rapid development of a diverse range of evolutionary algorithms (EAs). Most of the current state-of-the-art algorithm development is driven by the benchmarking problems developed as part of the evolutionary computation literature to test their performance. The result is that most of algorithms are specialists to these problems with a bias towards high convergence over diversity. They exhibit high performance on this range of problems, as can be seen in multiple competitive benchmarks \cite{Zhang2009b, Liu2017, Jiang2017} and is summarised in Table \ref{Tab:Lit_rev:GAs_test_problems_uncons} for unconstrained cases and in Table \ref{Tab:Lit_rev:GAs_test_problems_cons} for constrained and imbalanced problems; but when they are applied to practical applications they can exhibit poor performance, \cite{Grudniewski2019, Sobey2019, Wang2018, Mutlu2017}. This is supported by the lack of uptake of many of the algorithms with top performance in the evolutionary computation benchmarking exercises in the practical literature and the prevalence of NSGA-II in many fields; this is despite its original development dating back to 2002 and many newer solvers having been developed since then. It is the only current algorithm that has been shown to have high performance across the range of current benchmarking problems, shown in Tables \ref{Tab:Lit_rev:GAs_test_problems_uncons} - \ref{Tab:Lit_rev:GAs_test_problems_cons}, where the red boxes indicate a lack of testing on a given problem set.

	\begin{table}[!b]
	\caption{Summary of the literature on the unconstrained benchmarking sets used to test genetic algorithms.}
	\label{Tab:Lit_rev:GAs_test_problems_uncons}
	\begin{center}
		\begin{scriptsize}
			\begin{tabular}{>{\centering}m{1.5cm}|>{\centering}m{2cm}|>{\centering}m{2cm}|>{\centering}m{2cm}|>{\centering}m{2cm}}
				\hline
				& \multicolumn{4}{c}{\textbf{Test set}}                                 \tabularnewline \cline{2-5}
				\multirow{-2}{*}{\textbf{Algorithm}} & \textbf{ZDT}                      & \textbf{DTLZ}                                       & \textbf{WFG}               & \textbf{UF}                                    \tabularnewline \hline \hline
				\textbf{NSGA-II}                     & \cite{Deb2002, Lin2016}           & \cite{Deb2001, Seada2015, Lin2016}                  & \cite{Huband2005, Lin2016} & \cite{Zhang2009b, Lin2016}                         \tabularnewline \hline
				\textbf{NSGA-III}                    & \cite{Seada2015, Li2016}          & \cite{Deb2014, Seada2015, Li2016, Liu2019}          & \cite{Deb2014, Li2016}     & \cite{Li2016, Liu2019}                      \tabularnewline \hline
				\textbf{U-NSGA-III}                  & \cite{Seada2015}                  & \cite{Seada2015, Liu2019}                           & \cellcolor{redbox}   & \cite{Liu2019}                               \tabularnewline \hline
				\textbf{GAME}                        & \cellcolor{redbox}          & \cellcolor{redbox}                            & \cellcolor{redbox}   & \cellcolor{redbox}                    \tabularnewline \hline
				\textbf{aGAME}                       & \cellcolor{redbox}          & \cellcolor{redbox}                            & \cellcolor{redbox}   & \cellcolor{redbox}                    \tabularnewline \hline
				\textbf{MOEA/D}                      & \cite{Zhang2007, Lin2016, Li2016} & \cite{Deb2014, Zhang2007, Lin2016, Li2016, Liu2019} & \cite{Lin2016, Li2016}     & \cite{Zhang2009b, Lin2016, Li2016, Liu2019}    \tabularnewline \hline
				\textbf{MOEA/D-DE}                   & \cellcolor{redbox}          & \cellcolor{redbox}                            & \cellcolor{redbox}   & \cite{Chen2009}                              \tabularnewline \hline
				\textbf{MOEA/D-M2M}                  & \cellcolor{redbox}          & \cellcolor{redbox}                            & \cite{Jiang2018}           & \cite{Jiang2018}                             \tabularnewline \hline
				\textbf{MOEA/D-PSF}                  & \cellcolor{redbox}          & \cellcolor{redbox}                            & \cite{Jiang2018}           & \cite{Jiang2018}                             \tabularnewline \hline
				\textbf{MOEA/D-MSF}                  & \cellcolor{redbox}          & \cellcolor{redbox}                            & \cite{Jiang2018}           & \cite{Jiang2018}                             \tabularnewline \hline
				\textbf{LiuLi}                       & \cellcolor{redbox}          & \cellcolor{redbox}                            & \cellcolor{redbox}   & \cite{Zhang2009b}                           \tabularnewline \hline
				\textbf{DMOEA-DD}                    & \cellcolor{redbox}          & \cellcolor{redbox}                            & \cellcolor{redbox}   & \cite{Zhang2009b}                            \tabularnewline \hline
				\textbf{COEA}                        & \cite{Goh2009}                    & \cite{Goh2009}                                      & \cellcolor{redbox}   & \cellcolor{redbox}                     \tabularnewline \hline
				\textbf{HEIA}                        & \cite{Lin2016}                    & \cite{Lin2016}                                      & \cite{Lin2016}             & \cite{Lin2016}                              \tabularnewline \hline
				\textbf{BCE}                         & \cite{Li2016}                     & \cite{Li2016}                                       & \cite{Li2016}              & \cite{Li2016}                                \tabularnewline \hline
				\textbf{MTS}                         & \cite{Tseng2007}                  & \cite{Tseng2007}                                    & \cite{Tseng2007}           & \cite{Zhang2009b}                           \tabularnewline \hline
			\end{tabular}
		\end{scriptsize}
	\end{center}
\end{table}

\begin{table}[!t]
	\caption{Summary of the literature on the constrained and imbalanced benchmarking sets used to test genetic algorithms.}
	\label{Tab:Lit_rev:GAs_test_problems_cons}
	\begin{center}
		\begin{scriptsize}
			\begin{tabular}{>{\centering}m{1.5cm}||>{\centering}m{1.5cm}|>{\centering}m{1.5cm}|>{\centering}m{1.cm}|>{\centering}m{1.1cm}||>{\centering}m{1.2cm}|>{\centering}m{1.1cm}}
				\hline
				& \multicolumn{6}{c}{\textbf{Test set}}                                                                                                             \tabularnewline \cline{2-7}
				& \multicolumn{4}{c||}{\textbf{Constrained}}                                                                        & \multicolumn{2}{c}{\textbf{Imbalanced}}              \tabularnewline \cline{2-7}
				\multirow{-3}{*}{\textbf{Algorithm}}    & \textbf{CF}                  & \textbf{DTLZ}              & \textbf{IMB}             & \textbf{DAS-CMOP}        & \textbf{MOP}              & \textbf{IMB}             \tabularnewline \hline \hline
				\textbf{NSGA-II}                    & \cite{Zhang2009b}            & \cite{Deb2001, Seada2015}  & \cite{Liu2017}           & \cite{Fan2019}           & \cite{Liu2014}            & \cite{Liu2017}           \tabularnewline \hline
				\textbf{NSGA-III}                                     & \cellcolor{redbox}     & \cite{Jain2014, Seada2015} & \cellcolor{redbox} & \cite{Fan2019}           & \cellcolor{redbox}  & \cellcolor{redbox} \tabularnewline \hline
				\textbf{U-NSGA-III}                                   & \cellcolor{redbox}     & \cite{Seada2015}           & \cellcolor{redbox} & \cellcolor{redbox} & \cellcolor{redbox}  & \cellcolor{redbox} \tabularnewline \hline
				\textbf{GAME}                      & \cite{Abdou2012}             & \cellcolor{redbox}   & \cellcolor{redbox} & \cellcolor{redbox} & \cellcolor{redbox}  & \cellcolor{redbox} \tabularnewline \hline
				\textbf{aGAME}                             & \cite{Abdou2012a}            & \cellcolor{redbox}   & \cellcolor{redbox} & \cellcolor{redbox} & \cellcolor{redbox}  & \cellcolor{redbox} \tabularnewline \hline
				\textbf{MOEA/D}                  & \cellcolor{redbox}     & \cite{Jain2014}            & \cite{Liu2017}           & \cite{Fan2019}           & \cellcolor{redbox}  & \cite{Liu2017}           \tabularnewline \hline
				\textbf{MOEA/D-DE}                    & \cite{Chen2009}              & \cellcolor{redbox}   & \cellcolor{redbox} & \cellcolor{redbox} & \cite{Liu2014}            & \cellcolor{redbox} \tabularnewline \hline
				\textbf{MOEA/D-M2M}                    & \cellcolor{redbox}     & \cellcolor{redbox}   & \cellcolor{redbox} & \cellcolor{redbox} & \cite{Liu2014, Jiang2018} & \cellcolor{redbox} \tabularnewline \hline
				\textbf{MOEA/D-PSF}                             & \cellcolor{redbox}     & \cellcolor{redbox}   & \cellcolor{redbox} & \cellcolor{redbox} & \cite{Jiang2018}          & \cellcolor{redbox} \tabularnewline \hline
				\textbf{MOEA/D-MSF}                         & \cellcolor{redbox}     & \cellcolor{redbox}   & \cellcolor{redbox} & \cellcolor{redbox} & \cite{Jiang2018}          & \cellcolor{redbox} \tabularnewline \hline
				\textbf{LiuLi}                           & \cite{Zhang2009b, Abdou2012} & \cellcolor{redbox}   & \cellcolor{redbox} & \cellcolor{redbox} & \cellcolor{redbox}  & \cellcolor{redbox} \tabularnewline \hline
				\textbf{DMOEA-DD}                          & \cite{Zhang2009b, Abdou2012} & \cellcolor{redbox}   & \cellcolor{redbox} & \cellcolor{redbox} & \cellcolor{redbox}  & \cellcolor{redbox} \tabularnewline \hline
				\textbf{COEA}                        & \cellcolor{redbox}     & \cellcolor{redbox}   & \cellcolor{redbox} & \cellcolor{redbox} & \cellcolor{redbox}  & \cellcolor{redbox} \tabularnewline \hline
				\textbf{HEIA}                        & \cellcolor{redbox}     & \cellcolor{redbox}   & \cellcolor{redbox} & \cellcolor{redbox} & \cellcolor{redbox}  & \cellcolor{redbox} \tabularnewline \hline
				\textbf{BCE}                           & \cellcolor{redbox}     & \cellcolor{redbox}   & \cellcolor{redbox} & \cellcolor{redbox} & \cellcolor{redbox}  & \cellcolor{redbox} \tabularnewline \hline
				\textbf{MTS}                            & \cite{Zhang2009b, Abdou2012} & \cellcolor{redbox}   & \cellcolor{redbox} & \cellcolor{redbox} & \cellcolor{redbox}  & \cellcolor{redbox} \tabularnewline \hline
			\end{tabular}
		\end{scriptsize}
	\end{center}
\end{table}

There are a number of factors for why these new algorithms may not have found success in the applied literature. Firstly, in many practical applications the characteristics of the search and objective spaces are often not well understood prior to performing the optimisation. Therefore, it is not possible to select the most appropriate solvers in advance. Secondly, the benchmarking problems commonly utilised in the evolutionary literature are dominated by one main characteristic, whereas real-world problems may contain multiple dominant features. Lastly, the current state-of-the-art focuses on unconstrained continuous problems, with few efficient algorithms specialised for problems with other characteristic types, whilst most of real-world applications tend to have discontinuous, multi-modal and imbalanced search spaces. This is reflected in Table \ref{Tab:Lit_rev:GAs_test_problems_uncons} and in Table \ref{Tab:Lit_rev:GAs_test_problems_cons} where it is possible to see a trend towards limited benchmarking during the development of an algorithm, although many of these test problems were not available during the development of many of these algorithms, and only the top performing algorithms eventually being tested over a large range of test problems. There is a prevalence towards testing on unconstrained problems, with all but the GAME and aGAME tested on these problems, with more limited testing on the constrained and imbalanced problems.

It is known that convergence first algorithms struggle on a range of problems as they do not explore the entire decision variable and objective spaces due to their lack of diversity \cite{Liu2017, Liu2014}. This is supported by investigations on practical problems where algorithms with better diversity preservation show better performance as the complexity of the problem increases \cite{Grudniewski2019, Sobey2019}. This partly explains the success of NSGA-II, which has excellent diversity retention and can be classed as a general solver. Despite this, most of the currently developed evolutionary algorithms utilise a "convergence first, diversity second" approach \cite{Liu2017} where the mechanisms that promote convergence are preferred, while diversity is obtained using secondary methods such as crowding distance calculations \cite{Deb2002}, external archive refining \cite{Zitzler2001}, problem decomposition \cite{Zhang2007, Liu2009a, Liu2009b} or indicator-based solution selection \cite{Zitzler2004,Beume2007,Priester2013}. 

A recent methodology developed by Grudniewski and Sobey \cite{Sobey2018}, \cite{Grudniewski2018}, Multi-Level Selection Genetic Algorithm (MLSGA), introduces a Multi-Level Selection (MLS) mechanism, via sub-populations called collectives. The approach is based on distinct mechanisms at the individual and collective level, where the collective level utilises reproduction of sub-groups based on selection and elimination of collectives. This reproduction is based on distinctive fitness definitions, allowing information exchange between groups and ensuring that collectives with different fitness evaluations focus their search on different parts of the search space promoting a "diversity first, convergence second" approach. The diversity is generated by the unique region-based search, which is enhanced by separating the fitness function so that each level focuses its reproduction on different parts of the objective space. This approach has been shown to improve the diversity of the search and does not require extensive a priori knowledge about the problem. 

In parallel to the development of MLSGA, a new range of co-evolutionary algorithms, such as HEIA and BCE, have been shown to have strong general performance. They improve the generality of the search by merging multiple reproduction and/or improvement mechanisms which increases the chance that an effective approach will be available or by allowing the algorithm to self-adapt to the optimised problem during the process \cite{ Potter1994, DeJong2006}. However, in most cases the self-adapting algorithms have either high computational complexity or are sensitive to predefined hyperparameters \cite{Zhou2011}, making them impractical for real problems and therefore these approaches are not considered. Implementing these approaches without these complex mechanisms should allow a more general approach, with high diversity, without substantially reduced convergence. 

To provide an algorithm that can be used practically without a priori knowledge, this paper creates a novel co-evolutionary approach that works at the collective level rather than between individuals. In this case the different collectives utilise separate reproduction mechanisms, and the information exchange is performed via the collective-level reproduction mechanisms. Due to the lack of characterisation of complimentary mechanisms in co-evolutionary algorithms, different state-of-the-art Evolutionary Algorithms are implemented as the individual-level reproduction mechanisms of cMLSGA: Unified Non-dominated Sorting Genetic Algorithm (U-NSGA-III) \cite{Seada2015}, Indicator Based Evolutionary Algorithm (IBEA) \cite{Zitzler2004}, Multi-Objective Evolutionary Algorithm based on Decomposition (MOEA/D) \cite{Zhang2007}, Bi-Criterion Evolutionary Algorithm (BCE) \cite{Li2016}, Hybrid Evolutionary Immune Algorithm (HEIA) \cite{Lin2016}, Multiple-Trajectory Search (MTS) \cite{Tseng2009} and the updated MOEA/D variants with different Scalarization Functions MOEA/D-PSF and MOEA/D-MSF \cite{Jiang2018}. The resulting algorithms are tested on a number of multi-objective test functions \cite{Zhang2009, Zitzler2000,Huband2005,Liu2014, Liu2017}, and the top performing variants are selected and compared to the current state-of-the-art. 

The remainder of this paper is organised as follows. In section II, the mechanisms are compared to the related co-evolutionary algorithm literature. In section III the proposed methodology and the benchmarking exercise are described in detail. Section IV presents the performance of cMLSGA, with a comparison to the current state-of-the-art and is concluded in Section V.  			

\section{Co-evolutionary Genetic Algorithms}
Co-evolution is used to explain complex evolutionary dependencies between various groups of organisms \cite{Nusimer2017} such as: symbiosis \cite{Currie2003}, coadaptation \cite{Potter2000}, host-parasite \cite{Anderson1982} and hunter-prey \cite{Downes1998} relations. The term co-evolution is first introduced by Ehrlich and Raven in 1964 \cite{Ehrlich1964} to describe the coexistence of plants and butterflies where one group cannot survive without the other. However, the origins of this theory are older and they are clearly described by Darwin \cite{Darwin1859} when documenting the interactions between plants and insects. Co-evolution can occur in two forms: cooperation \cite{Potter1994}, where organisms coexist and “support” each other in various tasks or are mutually dependent for further benefits; or competition \cite{Hill1990}, where an “arms race” occurs between species as only the “strongest” may survive in the given environment.

The idea of introducing co-evolution into Evolutionary Algorithms is first presented by Potter and De Jong in 1994 \cite{Potter1994}. In the co-evolutionary approach multiple populations of species of individuals coexist and evolve in parallel, potentially utilising distinct reproduction mechanisms, with data exchange introduced between them. The sub-populations can operate on the same search space \cite{DeJong2006} or can be divided into several regions by problem decomposition and additional separation mechanisms \cite{Jia2018}. The form of data exchange between groups depends on the type of co-evolution utilised: competitive or cooperative. In cooperative co-evolution the genetic information, usually the decision variables, is shared by different species to form a valid solution when the problem is decomposed \cite{Potter1994}; or different sub-populations may cooperate to form the Pareto objective front with different subpopulations focusing on different regions \cite{CoelloCoello2003}. In competitive algorithms different groups compete in the creation of new populations \cite{Lin2016} or sub-populations \cite{Goh2009}, where fitter sub-populations gain a wider proportion of children in the next generation, or via an “arms race” where losing subpopulations try to counter the winning ones by using a faster evolution approach \cite{Rosin1997}.

In the evolutionary computation literature, multiple algorithms can be found which are inspired by co-evolution. These can be separated into two groups: approaches where all of the sub-populations are subject to the same individual reproduction strategies or those where distinct mechanisms are used to provide different performances. The algorithms where the individuals are subjected to the same reproduction strategies use other means to increase the diversity while maintaining convergence, such as the “species” approach, where individuals have different “traits” assigned that determine their selection process such as sex \cite{Raghuwanshi2006} or have a focus on different variables \cite{Li2003}. Otherwise problem decomposition\cite{Goh2009} is used to split the complex problem into sub-problems of a lower order. However, no increase in generality has been demonstrated for these methods. In the second approach, each group of individuals utilises distinct reproduction strategies, where each additional mechanism decreases the risk that all of the strategies will be ineffective on that particular problem at the same time, leading to a more general approach. Here only the algorithms of the second group are considered, due to the interest in improving the general performance of MLSGA, and the top performing algorithms are reviewed. 

There are two recent examples of algorithms using distinct reproduction strategies, BCE and HEIA. In the BCE \cite{Li2016}, sub-populations operate on the same search spaces and individuals for each group are selected at each generation based on two distinct fitness indicators: a Pareto-based criterion (PC) and a Non-Pareto-based criterion (NPC). In the Pareto-based criterion, standard Pareto dominance is utilised based on the objective functions, which rewards convergence, whereas the Non-Pareto based selection uses an additional indicator that rewards diversity of solutions, based on Hyper-Volumes. Therefore, different sub-population evaluation schemes are utilised, rather than distinct evolutionary strategies. This leads to an overall improvement in diversity for the entire population, especially on many-objective cases and problems with irregular search spaces and variable linkages. Additionally, it has been shown that different methodologies can be utilised for both the Pareto and Non-Pareto criteria searches, without extensive parameter tuning. However, the search is still convergence dominated and shows decreasing performance on practical problems of increasing complexity \cite{Sobey2019}. This shows the potential for algorithms that combine a range of mechanisms in the co-evolutionary search. A similar approach has been utilised in HEIA \cite{Lin2016}, but in this case two distinct evolutionary computation methods are used, Immune Algorithm and a Genetic Algorithm, instead of separate quality indicators. In this method the best individuals are moved to a shared pool at each generation and the sub-populations are recreated from the pool of individuals in a cloning process. However, despite the potential of combining multiple distinct mechanisms, shown by BCE and HEIA, there is a lack of comprehensive documentation that would explain which combinations of mechanisms should be chosen for a particular problem in order to achieve a high performance. Therefore, the mechanisms appear to have been chosen arbitrarily, with the potential for improvements with a clearer documentation of how different mechanisms pair. 

A similar approach to the co-evolutionary algorithms can be found in the island-based methods \cite{Kurdi2016, Whitley1999} and certain  decomposition-based GAs, such as M2M \cite{Liu2014} and DMOEA-DD \cite{Liu2009a}, which use sub-populations in a similar way to the  co-evolutionary approaches. For island-based methods, the sub-population based approach is introduced with information exchange in the form of migration of individuals. However, due to a lack of proper diversity preservation mechanisms, island-based methods are predominantly utilised for single-objective optimisation and specific problems, such as scheduling \cite{Kurdi2016}. In both M2M and DMOEA-DD, multiple sub-populations are also maintained and evolve in parallel. In both DMOEA-DD and M2M the mechanisms are based on a forced problem decomposition of the search spaces, which means that the individuals are assigned to a specific sub-population dependant on their current positions. However, the combination of assigning sub-populations and the problem decomposition requires an extensive a priori knowledge about both the search and objective spaces \cite{Trivedi2016}. Many problem decomposition methods are shown to be ineffective on problems with highly discontinuous areas, as the weight vectors cannot operate freely around the infeasible regions \cite{Trivedi2016}, and therefore application of these methods to real-life problems is limited. 

The co-evolutionary approaches are therefore preferred as they provide an additional diversity and generality of search, with minimal loss in convergence and their performance is robust to the hyperparameter selection. However, none of these current approaches provide a fitness definition specific to the sub-population or reproduction mechanisms for these sub-populations to allow them to evolve in combination with the individuals within them.

\section{cMLSGA}
\subsection{Multi-Level Selection Genetic Algorithm}

A detailed explanation of the mechanisms and principles of working of MLSGA can be found in Sobey and Grudniewski \cite{Sobey2018} and its extension to use other algorithms at the individual level to improve convergence can be found in Grudniewski and Sobey \cite{Grudniewski2018}. However, a short review of the mechanisms is provided here for clarity. The high performance of MLSGA is based on two novel mechanisms: collective reproduction, which creates an additional selection pressure and enhances the convergence rate; and the separation of the fitness between levels, which encourages each collective to explore different areas and greatly increases the diversity of the obtained solutions. In MLSGA the whole population is separated into a number of sub-groups, called collectives, each operating on the same search and objective spaces. These collectives evolve in parallel using the same reproduction mechanism, but after a predefined, usually small, number of generations the worst collective is eliminated, based on its collective fitness value, and repopulated by individuals from the other collectives. Importantly, the collective has a specific fitness definition which can be different to the individuals inside of it. This is inspired by the evolutionary literature which demonstrates a split in how the fitness should be defined for each collective with two distinct options: MLS1 and MLS2. The MLS1 fitness is defined as the aggregate of the objective functions for the individuals in the collective and in MLSGA this focuses the search towards the centre of the Pareto Front; MLS2 is defined as there being a different fitness between the individuals and the collective. Therefore, in MLSGA, using a bi-objective problem as an example, the collectives have one objective assigned to them and the individuals the other, with the result that the focus is on one extreme region of the Pareto front. By creating a reverse of this, MLS2R, the remaining collectives focus their search on the other extreme values. Therefore, each collective is "primed" to explore different regions of the objective space, leading to a sub-regional search strategy. As each collective’s survival and convergence is dependent on their ability to find diverse regions of the objective space, MLSGA follows a “diversity first, convergence second” principle.  

\subsection{Novel co-evolutionary approach}
\begin{figure}[!t]
	\begin{center}
		\includegraphics[width=0.55\textwidth]{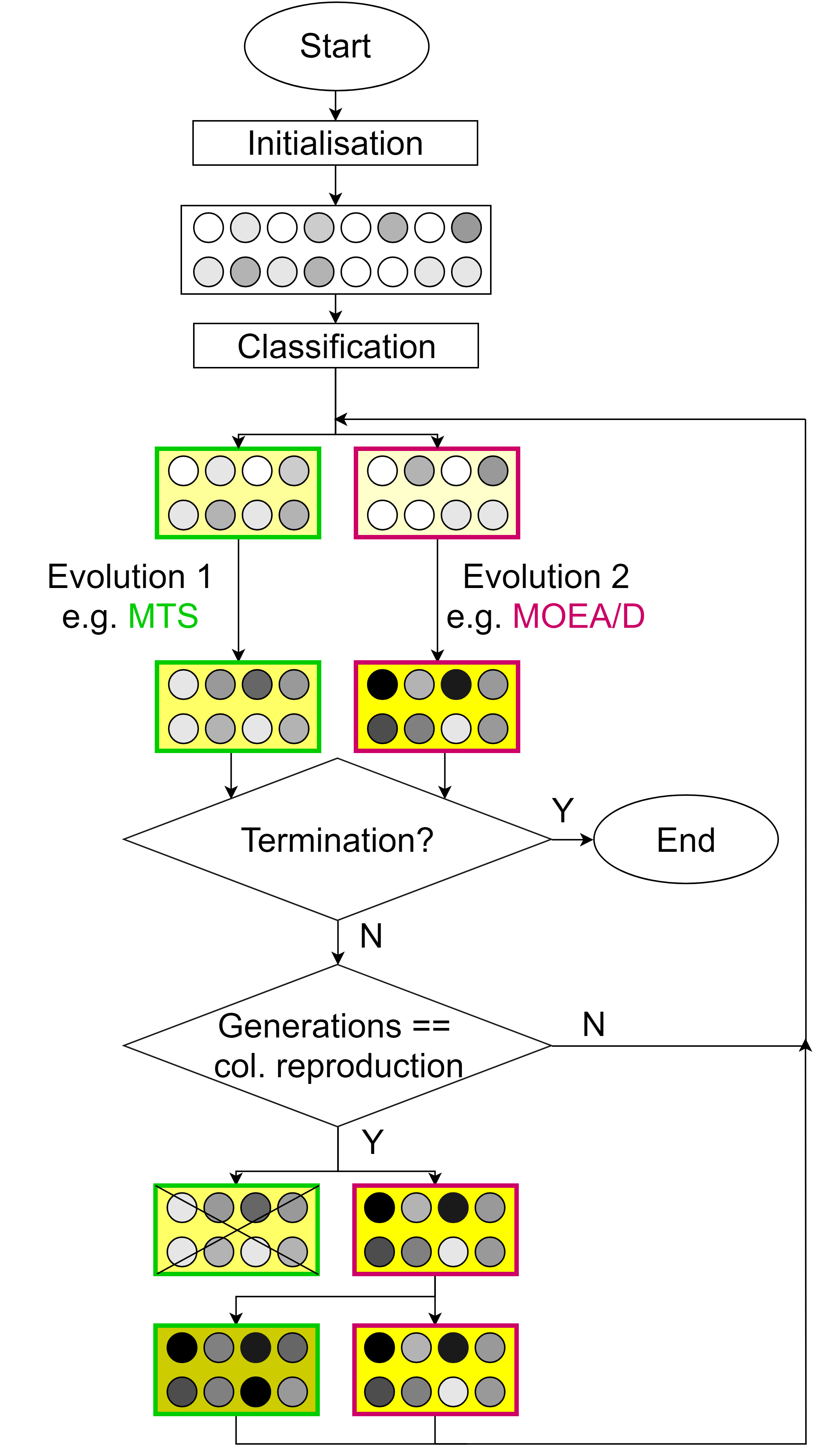}
		\caption{cMLSGA methodology. Only two collectives are shown to illustrate how two distinct evolutionary strategies work in combination; in the results eight collectives are used.}
		\label{fig:F2_cMLSGA}
		
	\end{center}
\end{figure}

In the previously developed hybrid approach each collective utilises the same mechanisms for the individual-level reproduction, taken from the current state-of-the-art. In this paper a novel method is suggested which utilises the multi-level nature of MLSGA to develop a distinct co-evolutionary approach that is competitive through its collective level reproduction mechanisms. Different collectives utilise different mechanisms for individual offspring creation and the information exchange is less regular than in other approaches, encouraging collectives to act independently. All of the original mechanisms, such as classification and collective reproduction, are copied from the previous hybrid algorithms \cite{Grudniewski2018}, and the same principle applies to the reproduced state-of-the-art algorithms. The methodology for cMLSGA is presented in Fig. \ref{fig:F2_cMLSGA}, where darker circles indicate fitter individuals and darker yellow rectangles indicate higher collective fitness. The new algorithm, competitor algorithms and benchmarking are coded in C++\footnote{The code is available online with detailed instructions at: https://www.bitbucket.org/Pag1c18}.

Similarly to MLSGA, the randomly generated initial population is first classified into a predefined number of collectives using a Support Vector Machine, based on the decision variable space. Importantly, the classification step occurs only once and is not repeated over the course of the optimisation. In this case, the multi-class classification SVM with C parameter and linear function, called C-SVC, is used. The utilised code has been taken from the LIBSVM open library \cite{Chang2011} and the training parameters remains the same as in the publication.  Each collective has a separate fitness definition assigned, as explained in \cite{Grudniewski2018}. 

In cMLSGA there is an additional step where half of the collectives utilise the selected "Evolutionary Strategy 1" (ES1) subroutine, and the other half the ”Evolutionary Strategy 2” (ES2). The type of evolution is assigned randomly to the collectives during the classification step and does not change over the run, at any time half of the collectives are using ES1 and the other half use ES2. Next, the individuals in each sub-population evolve separately using the assigned strategy. After a predefined number of generations, the collective with the worst fitness value is eliminated, with all of the individuals inside of it, and is repopulated by copying the best individuals from among the rest of the collectives. 

After the collective reproduction step the offspring collective inherits the individual reproduction methodology from the eliminated collective, demonstrated further in Fig. \ref{fig:F2_cMLSGA} where the left collective retains MTS and the right MOEA/D. Therefore, cMLSGA utilises the multi-level approach to generate a distinct competitive co-evolutionary approach but this occurs via elimination and repopulation of one group by other sub-populations, rather than migration of individuals \cite{Li2016}, competition in each group \cite{Goh2009} or recreation of all groups \cite{Lin2016}. In addition, unlike in other co-evolutionary GAs, the sub-populations are not allowed to exchange information every generation, but only during the collective reproduction steps. This leads to a high diversity search and a lack of bias towards specific problem types.

\subsection{Computational complexity and constraint handling}

In cMLSGA, no computationally expensive mechanisms are added to the original MLSGA. Therefore, the computational complexity of one generation of cMLSGA is the same as MLSGA, and is bounded by $O(mN^2)$, where $m$ is the number of objectives and $N$ is the population size; or $C$, which is the complexity of the most complex embedded algorithm. More details can be found in \cite{Grudniewski2018}. The constraint-domination principle is adopted from NSGA-II \cite{Deb2002}.	This applies whenever two individuals are compared, similarly to the original MLSGA no constraint handling is introduced for the collective comparison and selection.
\pagebreak
\section{Benchmarking cMLSGA performance}
\subsection{Test problems}
In order to investigate the performance over cases with different dominate characteristics the proposed methodology is verified over 60 state-of-the-art bi-objective and 40 three-objective test functions \cite{Zhang2009, Zitzler2000, Huband2005, Liu2014, Liu2017, Fan2019}. These problems are divided into twelve categories for the 2-objective problems and nine for the 3-objective problems, with the justification for the selection given in Tables \ref{Tab:Test_set_uncons} and \ref{Tab:Many_TestSet}. Due to the variety of the problems in the test set two and three objective problems are preferred as they allow a simpler interpretation of the results, without issues of scaling, while providing enough complexity to resemble a number of real-world problems. However, like most benchmarking problems in evolutionary computation, they are dominated by a single characteristic, which is less common in real-world problems.

\begin{center}
	\begin{footnotesize}
		\begin{longtable}{>{\centering}m{2.5cm}|>{\centering}m{2.5cm}|>{\centering}m{0.2cm}|>{\centering}m{5.25cm}}
			\caption{Summary of the utilised two-objective test set }
			\label{Tab:Test_set_uncons}
			\tabularnewline \hline
			\textbf{Category} &\textbf{Problem} & \textbf{d} & \textbf{Additional properties}
			\tabularnewline \hhline{====} \endfirsthead
			\caption{Summary of the utilised two-objective test set (continued) }
			\tabularnewline \hline
			\textbf{Category} &\textbf{Problem} & \textbf{d} & \textbf{Additional properties}
			\tabularnewline  \hhline{====} \endhead
			\multicolumn{4}{c}{\textbf{Unconstrained}}   
			\tabularnewline  \hhline{====}
			\multirow{5}{2.6cm}{\textbf{I. Simple}} 
			& ZDT1 & 30 & Convex
			\tabularnewline \cline{2-4}	
			& ZDT2 & 30 &  Concave
			\tabularnewline \cline{2-4}	
			& ZDT3 & 30 & 	Discontinuous
			\tabularnewline \cline{2-4}	
			& ZDT4 & 10 & 	Multimodal, Convex
			\tabularnewline \cline{2-4}
			& ZDT6 & 10 & 	Multimodal, Biased, Concave
			\tabularnewline \hline
			
			\multirow{3}{2.6cm}{\textbf{II. Convex}} 
			& UF1 & 30 & Complex PS	
			\tabularnewline \cline{2-4}	
			& UF2 & 30 & Complex PS
			\tabularnewline \cline{2-4}
			& UF3 & 30 & Complex PS	
			\tabularnewline \hline
			
			\multirow{7}{2.6cm}{\textbf{III. Concave}} 
			& UF4 & 30 & Complex PS	
			\tabularnewline \cline{2-4}	
			& WFG4 & 22 & 	Multimodal
			\tabularnewline \cline{2-4}	
			& WFG5 & 22 & 	Deceptive
			\tabularnewline \cline{2-4}	
			& WFG6 & 22 & 	Non-separable
			\tabularnewline \cline{2-4}	
			& WFG7 & 22 & 	Biased
			\tabularnewline \cline{2-4}	
			& WFG8 & 22 & 	Biased, Non-separable
			\tabularnewline \cline{2-4}	
			& WFG9 & 22 & 	Biased, Non-separable, Deceptive	
			\tabularnewline \hline
			
			\multirow{3}{2.6cm}{\textbf{IV. Linear/Mixed}}
			& UF7 & 30 & Complex PS, Linear		
			\tabularnewline \cline{2-4}
			& WFG1 & 22 & Biased, Mixed	
			\tabularnewline \cline{2-4}
			& WFG3 & 22 & Non-separable, Degenerated, Linear
			\tabularnewline \hline
			
			\multirow{4}{2.6cm}{\textbf{V. Discontinuous}}
			& UF5 & 30 & Linear, Distinct points, Complex PS	
			\tabularnewline \cline{2-4}
			& UF6 & 30 & Complex PS
			\tabularnewline \cline{2-4}					
			& WFG2 & 22 & Convex, Non-Separable
			\tabularnewline \cline{2-4}
			& MOP4 & 10 & Discontinuous	
			\tabularnewline \hline
			
			\multirow{10}{2.6cm}{\textbf{VI. Imbalanced}} 
			& MOP1 & 10 & Convex						
			\tabularnewline \cline{2-4}	
			& MOP2 & 10 & Convex						
			\tabularnewline \cline{2-4}
			& MOP3 & 10 & Concave						
			\tabularnewline \cline{2-4}					
			& MOP5 & 10 & Convex						
			\tabularnewline \cline{2-4}	
			& IMB1 & 10 & Convex						
			\tabularnewline \cline{2-4}
			& IMB2 & 10 & Linear						
			\tabularnewline \cline{2-4}
			& IMB3 & 10 & Concave						
			\tabularnewline \cline{2-4}
			& IMB7 & 10 & Convex, Non-separable						
			\tabularnewline \cline{2-4}
			& IMB8 & 10 & Linear, Non-separable						
			\tabularnewline \cline{2-4}
			& IMB9 & 10 & Concave, Non-separable							
			\tabularnewline  \hhline{====}
			\multicolumn{4}{c}{\textbf{Constrained}}                                                                      \tabularnewline  \hhline{====}
			
			\multirow{3}{2.6cm}{\textbf{VII. Discontinuous}} 
			& CF1 & 10 & Linear, Complex PS, Distinct points
			\tabularnewline \cline{2-4}
			& CF2 & 10 & Convex, Complex PS
			\tabularnewline \cline{2-4}
			& CF3 & 10 & Concave, Complex PS
			\tabularnewline \hline
			
			\multirow{3}{2.6cm}{\textbf{VIII. Continuous}} & CF4 & 10 & Linear, Complex PS
			\tabularnewline \cline{2-4}
			& CF5 & 10 & Linear, Complex PS
			\tabularnewline \cline{2-4}
			& CF6 & 10 & Mixed, Complex PS
			\tabularnewline \cline{2-4}
			& CF7 & 10 & Mixed, Complex PS
			\tabularnewline \hline	
			
			\multirow{3}{2.6cm}{\textbf{IX. Imbalanced}} & IMB11 & 10 & Convex						
			\tabularnewline \cline{2-4}
			& IMB12 & 10 & Linear					
			\tabularnewline \cline{2-4}
			& IMB13 & 10 & Concave	
			\tabularnewline \hline
			
			\multirow{6}{2.6cm}{\textbf{X. Diversity-hard}} & DAS-CMOP1(5) & 30 & Concave, Discontinuous						
			\tabularnewline \cline{2-4}
			& DAS-CMOP2(5) & 30 & Mixed, Continuous						
			\tabularnewline \cline{2-4}	
			& DAS-CMOP3(5) & 30 & Linear, Discontinuous, Multimodal						
			\tabularnewline \cline{2-4}
			& DAS-CMOP4(5) & 30 & Concave, Discontinuous						
			\tabularnewline \cline{2-4}
			& DAS-CMOP5(5) & 30 & Mixed, Discontinuous						
			\tabularnewline \cline{2-4}
			& DAS-CMOP6(5) & 30 & Distinct points, Degenerated
			\tabularnewline \hline
			\pagebreak
			\multirow{6}{2.6cm}{\textbf{XI. Feasibility-hard}} & DAS-CMOP1(6) & 30 & Concave, Discontinuous						
			\tabularnewline \cline{2-4}
			& DAS-CMOP2(6) & 30 & Mixed, Continuous						
			\tabularnewline \cline{2-4}	
			& DAS-CMOP3(6) & 30 & Linear, Discontinuous, Multimodal						
			\tabularnewline \cline{2-4}
			& DAS-CMOP4(6) & 30 & Concave, Discontinuous						
			\tabularnewline \cline{2-4}
			& DAS-CMOP5(6)) & 30 & Mixed, Discontinuous						
			\tabularnewline \cline{2-4}
			& DAS-CMOP6(6) & 30 & Distinct points, Degenerated
			\tabularnewline \hline
			
			\multirow{6}{2.6cm}{\textbf{XII.Convergence-hard}} & DAS-CMOP1(7) & 30 & Concave, Discontinuous						
			\tabularnewline \cline{2-4}
			& DAS-CMOP2(7) & 30 & Mixed, Continuous						
			\tabularnewline \cline{2-4}	
			& DAS-CMOP3(7)) & 30 & Linear, Discontinuous, Multimodal						
			\tabularnewline \cline{2-4}
			& DAS-CMOP4(7) & 30 & Concave, Discontinuous						
			\tabularnewline \cline{2-4}
			& DAS-CMOP5(7) & 30 & Mixed, Discontinuous						
			\tabularnewline \cline{2-4}
			& DAS-CMOP6(7) & 30 & Distinct points, Degenerated
			\tabularnewline \hhline{====}					
			
			\multicolumn{4}{c}{\textit{d denotes the number of decision variables.}}
			
		\end{longtable}
	
		\begin{longtable}{>{\centering}m{2.5cm}|>{\centering}m{2.5cm}|>{\centering}m{0.2cm}|>{\centering}m{5.25cm}}
			\caption{Summary of the utilised three-objective test set }
			\label{Tab:Many_test_set}
			\tabularnewline \hline
			\textbf{Category} &\textbf{Problem} & \textbf{d} & \textbf{Additional properties}
			\tabularnewline \hhline{====} \endfirsthead
			\caption{Summary of the utilised three-objective test set (continued) }
			\tabularnewline \hline
			\textbf{Category} &\textbf{Problem} & \textbf{d} & \textbf{Additional properties}
			\tabularnewline  \hhline{====} \endhead
				\multicolumn{4}{c}{\textbf{Unconstrained}}                                                                      \tabularnewline  \hhline{====}
				\multirow{13}{*}{\textbf{I. Concave}}        & DTLZ2  & 12        &                                    \tabularnewline   \cline{2-4}
				& DTLZ3    & 12        & Multimodal                         \tabularnewline   \cline{2-4}
				& DTLZ4   & 12        & Biased                             \tabularnewline   \cline{2-4} 
				& DTLZ5   & 12        & Degenerated                        \tabularnewline   \cline{2-4} 
				& DTLZ6   & 12        & Degenerated, Biased                \tabularnewline  \cline{2-4} 
				& UF8               & 30         & Complex PS                         \tabularnewline   \cline{2-4}
				& UF10              & 30         & Complex PS                         \tabularnewline  \cline{2-4} 
				& WFG4    & 24      & Multimodal                         \tabularnewline  \cline{2-4} 
				& WFG5    & 24      & Deceptive                          \tabularnewline   \cline{2-4}
				& WFG6    & 24      & Non-separable                      \tabularnewline   \cline{2-4}
				& WFG7    & 24      & Biased                             \tabularnewline   \cline{2-4}
				& WFG8    & 24      & Biased, Non-separable              \tabularnewline   \cline{2-4}
				& WFG9    & 24      & Biased, Non-separable, Deceptive   \tabularnewline  \hline 
				\multirow{3}{*}{\textbf{IV. Linear/Mixed}}   & DTLZ1   & 7        & Linear, Multimodal                 \tabularnewline \cline{2-4}  
				& WFG1    & 24      & Biased, Mixed                      \tabularnewline \cline{2-4}   
				& WFG3    & 24      & Non-separable, Degenerated, Linear \tabularnewline  \hline 
				\multirow{3}{*}{\textbf{V. Discontinuous}} & DTLZ7   & 22       & Mixed, Multimodal                  \tabularnewline   \cline{2-4}
				& UF9               & 30         & Complex PS                         \tabularnewline \cline{2-4} 
				& WFG2    & 24      & Convex, Non-Separable              \tabularnewline  \hline 
				\multirow{6}{*}{\textbf{VI. Imbalanced}}     & MOP6             & 10         & Linear                             \tabularnewline \cline{2-4}  
				& MOP7              & 10         & Concave                            \tabularnewline  \cline{2-4} 
				& IMB4              & 10         & Linear                             \tabularnewline   \cline{2-4}
				& IMB5              & 10         & Concave                            \tabularnewline   \cline{2-4}
				& IMB6              & 10         & Linear                             \tabularnewline   \cline{2-4}
				& IMB10             & 10         & Linear                             \tabularnewline  \hhline{====}
				
				\multicolumn{4}{c}{\textbf{Constrained}}                                                                        \tabularnewline  \hhline{====}
				\multirow{5}{*}{\textbf{VII. Discontinuous}}   & DTLZ8  & 30        & Mixed, Degenerated, Biased         \tabularnewline   \cline{2-4}
				& DTLZ9   & 30        & Concave, Degenerated               \tabularnewline  \cline{2-4} 
				& CF8              & 10         & Concave, Degenerated, Complex PS                \tabularnewline \cline{2-4}  
				& CF9               & 10         & Concave, Complex PS                             \tabularnewline   \cline{2-4}
				& CF10              & 10         & Concave, Complex PS                             \tabularnewline  \hline 
				\textbf{IX. Imbalanced}                      & IMB14             & 10         & Linear                             \tabularnewline  \hline
				
				\multirow{3}{2.6cm}{\textbf{X. Diversity-hard}} 
				& DAS-CMOP7(5) & 30 & Linear, Degenerated, Discontinuous					
				\tabularnewline \cline{2-4}
				& DAS-CMOP8(5) & 30 & Concave, Discontinuous						
				\tabularnewline \cline{2-4}	
				& DAS-CMOP9(5) & 30 &  Concave, Discontinuous, Biased
				\tabularnewline \hline  
				
				\multirow{3}{2.6cm}{\textbf{XI. Feasibility-hard}} 
				& DAS-CMOP7(6) & 30 & Linear, Degenerated, Discontinuous					
				\tabularnewline \cline{2-4}
				& DAS-CMOP8(6) & 30 & Concave, Discontinuous						
				\tabularnewline \cline{2-4}	
				& DAS-CMOP9(6) & 30 &  Concave, Discontinuous, Biased
				\tabularnewline \hline  
				
				\multirow{3}{2.6cm}{\textbf{XII.Convergence-hard}} 
				& DAS-CMOP7(7) & 30 & Linear, Degenerated, Discontinuous					
				\tabularnewline \cline{2-4}
				& DAS-CMOP8(7) & 30 & Concave, Discontinuous						
				\tabularnewline \cline{2-4}	
				& DAS-CMOP9(7) & 30 &  Concave, Discontinuous, Biased
				\tabularnewline \hhline{====}		
				
				\multicolumn{4}{c}{\textit{d denotes the number of decision variables. The same categories are utilised as for the two-objective cases.}} 
					
		\end{longtable}
	\end{footnotesize}
\end{center}
\subsection{Competitor algorithms and individual reproduction mechanisms}
In this study, a variety of reproduction mechanisms are replicated from the original papers and combined through the cMLSGA methodology: IBEA \cite{Zitzler2004} as the most commonly utilised indicator based GA; BCE \cite{Li2016} and HEIA \cite{Lin2016} as the most highly performing co-evolutionary approaches, demonstrating the difference in performance with the proposed methodology; MOEA/D-TCH \cite{Zhang2007} and  MOEA/D-(PSF/MSF) \cite{Jiang2018} as different variants of the top performing solver for unconstrained and imbalanced problems respectively; MTS \cite{Tseng2009} as a proficient constrained and unconstrained solver that utilises local search strategies instead of typical evolutionary mechanisms and U-NSGA-III \cite{Seada2015} as the universal two/many-objective variant of the most commonly utilised general-solver NSGA-II \cite{Deb2002}.

Each utilised mechanism is combined with every other mechanism in the co-evolutionary approach.  However, the combinations between U-NSGA-III, MOEA/D-TCH, MOEA/D-PSF and MOEA/D-MSF are discarded. This is because all of the MOEA/D variants are based on the same framework while U-NSGA-III and MOEA/D are both based on developing a constant Pareto front at each generation. The pre-benchmarks demonstrate that similar mechanisms result in no performance gains for co-evolutionary approaches, and the results for these variants are not reported. Therefore, 22 variants in total are evaluated with the results attached as a supplement to this paper.

The tests are performed over 30 separate runs, with 300,000 function evaluations for each run. The results are compared using two performance indicators: Inverted Generational Distance (IGD) and Hyper-Volume (HV), as they provide information on the convergence and diversity of the obtained solutions. IGD measures the average Euclidean distance between any point in a uniformly distributed Pareto front and the closest solution in the obtained set. Low scores for this metric emphasise convergence and uniformity of the points \cite{Zitzler2002}. HV is the measure of volume of the objective space between a predefined reference point and the obtained solutions, and therefore has a stronger focus on the diversity and edge points. 1,000,000 reference points are generated in this paper for the IGD calculation, and the HV is calculated according to While et al. \cite{While2012}, which provides the fastest and most widely used method for this calculation.

\subsection{Hyperparameters}

Different operational parameters: population size, number of collectives and collective reproduction delay, were tested and the best performing values are summarised in Table \ref{Tab:Bck_param}. However, it has been concluded that these parameters do not have significant impact on the performance. Between different groups of mechanisms, the only parameter that is different is the collective reproduction delay. The variance is caused by BCE, HEIA and IBEA developing fronts at different paces and therefore low values of collective elimination lead to premature removal of these underdeveloped fronts. MTS requires significantly more iterations per generation, as it utilises multiple local search methods, and therefore the elimination has to occur more frequently to have an impact on the search. The number of collectives is 8 in all cases, with 4 collectives using the first evolutionary algorithm and the other 4 using the second. All other GA specific parameters e.g. MOEA/D Neighbourhood size; are taken directly from the original publications to create a more realistic scenario where a priori knowledge cannot be utilised.

\begin{table}[!t]
	\caption{The GA parameters utilised for benchmarking}
	\label{Tab:Bck_param}
	\begin{center}
		\begin{footnotesize}
			\begin{tabular}{>{\centering}m{2.5cm}|>{\centering}m{0.9cm}|>{\centering}m{0.9cm}|>{\centering}m{0.9cm}|>{\centering}m{0.9cm}}
				\hline
				\textbf{Parameter} & \textbf{MTS} & \textbf{BCE} & \textbf{HEIA} & \textbf{IBEA} 
				\tabularnewline \hline \hline
				\textbf{Pop. Size} & \multicolumn{4}{c}{1000} 
				\tabularnewline \hline
				\textbf{Crossover rate} & \multicolumn{4}{c}{1} 
				\tabularnewline \hline
				\textbf{Mutation rate} & \multicolumn{4}{c}{0.08}
				\tabularnewline \hline
				\textbf{No. of collectives} & \multicolumn{4}{c}{8}
				\tabularnewline \hline
				\textbf{Collective reproduction delay} & 1 & 4 & \multicolumn{2}{c}{10}
				\tabularnewline \hline \hline
			\end{tabular}
		\end{footnotesize}
	\end{center}
\end{table}

\subsection{Determining the characteristics of complementary co-evolutionary mechanisms}

The performances of the selected cMLSGA variants are compared to the original algorithms with the data attached as a supplement. Due to the large quantity of data, only the most interesting 4 representative cases of the 22 methods trialled are described further in this section. IBEA\_BCE, which shows the highest improvement over the original algorithms; MOEA/D-MSF\_HEIA as the combination of a convergence-solver and a more general solver;   MOEA/D-MSF\_U-NSGA-III, as the combination of a convergence-solver and the top algorithm for problems with many objectives and MTS\_MOEA/D, which combines two convergence first methodologies: MTS which is a highly proficient constrained solver and MOEA/D, the top unconstrained solver. This is summarised in Table \ref{Tab:cMLSGA:Comp}. It can be observed that the cMLSGA approach does not improve the performance of the original algorithms in all cases and for the majority of them its performance is in between the effectiveness of both implemented methodologies. This shows that cMLSGA is often limited by the performance of the worst of the utilised strategies but leads to a more robust, "general solver", with a lower chance of poor performance.

\begin{table}[!b]
	\caption{Comparison of selected cMLSGA variants to the implemented algorithms according to IGD/HV indicator.}
	\label{Tab:cMLSGA:Comp}
	\begin{footnotesize}
		\begin{tabular}{>{\centering}m{2.5cm}|>{\centering}m{1.5cm}|>{\centering}m{1.6cm}|>{\centering}m{2.4cm}|>{\centering}m{2cm}}
			\hline
			\textbf{cMLSGA variant}   & \textbf{IBEA\_BCE} & \textbf{MOEA/D-MSF\_HEIA} & \textbf{MOEA/D-MSF \_U-NSGA-III} & \textbf{MTS\_MOEA/D} \tabularnewline \hline \hline
			\textbf{Better than both} & 67/62              & 38/40                     & 33/31                           & 29/23                \tabularnewline \hline
			\textbf{Better than one}  & 28/32              & 46/44                     & 57/52                           & 52/56                \tabularnewline \hline
			\textbf{Worse than both}  & 5/6                & 16/16                     & 10/17                           & 19/21                \tabularnewline \hline
		\end{tabular}
	\end{footnotesize}
\end{table}

Comparing the 22 variants of cMLSGA against the original algorithms the highest improvement is exhibited by the cMLSGA\footnotesize \_IBEA\_BCE \normalsize variant. In this case the performance is improved on 67 out of 100 problems for the IGD indicator and 62 cases for the HV indicator; it only exhibits worse performance in 5 and 6 cases for IGD and HV respectively. cMLSGA\footnotesize \_MOEA/D-MSF\_HEIA \normalsize is the best performer on the two-objective test set, where it is better than both methodologies in 38 cases for IGD and 40 cases for HV but worse in 16 cases for each indicator. Whereas cMLSGA\footnotesize \_MOEA/D-MSF\_U-NSGA-III \normalsize is the top performer for the three-objective problems. It performs better on 33 problems but is worse for 10 problems for IGD and is better for 31 problems and worse for 17 problems when evaluating using the HV metric. For the high convergence solver, cMLSGA\footnotesize \_MTS\_MOEA/D \normalsize better results are obtained in 29 and 23 cases for IGD and HV indicators but worse on 19 and 21 of them respectively. This indicates that if two methodologies with strong diversity mechanisms, such as IBEA and BCE, are used then cMLSGA is able to successfully combine the methodologies and preserve the diversity mechanisms, while providing stronger convergence. It supports the previous findings that the MLSGA mechanisms promote diversity \cite{Sobey2018}, but also indicates that if sufficient diversity is obtained the MLS-U variants focus the search, improving convergence. The reduced performance of the other two presented variants, which combine strong convergence based mechanisms, may be caused by the inability of MLSGA to properly maintain a search pattern during the collective elimination and reproduction steps, as indicated previously in \cite{Grudniewski2018}. In the case of the MTS and MOEA/D variants the performance of the cMLSGA approach resembles the results of MTS, especially for the IGD indicator. This is due to the local searches that require a significant number of iterations per generation, the MTS mechanisms "dominate" the search, and the secondary mechanisms, MLSGA and MOEA/D, do not have a sufficient influence. Therefore, if a method is combined with a significantly stronger convergence-first methodology than the other method, such as MTS, the weaker method will have negligible impact on the search. The result is that there is an improvement in the uniformity of the points, but no new areas of the Pareto optimal front are found and so diversity is not increased. A similar principle applies to the MOEA/D-MSF\_HEIA and MOEA/D-MSF\_U-NSGA-III variants, where the specialist convergence-based methodology is combined with a general diversity-based one. This indicates, that cMLSGA maintains distinct searches for both algorithms if two methodologies with contrasting mechanisms are combined, rather than one method outperforming the other. It can be concluded that the success in combining methodologies is subject to how similar the methods are and the poor performance of MOEA/D, MOEA/D-MSF and MOEA/D-PSF, compared to the original algorithms, shows that implementing methodologies that are too similar is not at all beneficial, due to the non-complementary nature of the searches, and that the diversity of the approaches is more important than their individual performances.

\subsection{Behaviour of cMLSGA: increased generality}

cMLSGA predominantly exhibits high performance on problems that require high diversity in comparison to the original algorithms. It operates significantly better on cases with discontinuous search spaces, categories V, VII and VIII, rather than continuous problems, and on diversity- and feasibility-hard functions, categories X and XI. The lower impact on the performance of the continuous functions is caused by the additional region searches provided by cMLSGA, which are not necessary as convergence is more important for these problems. Therefore, the presented methodology does not lead to performance improvements on the cases where the results from at least one of the original algorithms are close to the ideal Pareto optimal front. This is caused by there being little room for improvement, indicating that these mechanisms are well tuned to these problems and any changes reduce their effectiveness or that even more specialist mechanisms are required to improve the performance. Therefore, adding additional mechanisms that are created to enhance the overall diversity makes the search less focused and leads to worse overall performance if the algorithm is already able to obtain excellent results.  Unsurprisingly this indicates that cMLSGA will underperform convergence specialised GA methodologies on their preferred cases. This is demonstrated on the category XII problems, which are difficult convergence problems, with all of the cMLSGA variants performing relatively poorly. In contrast to the results on the convergence based problems, high performance is achieved on categories X-XI, which belong to the same test set \cite{Fan2019}, but focuses on providing hard diversity and feasibility problems.

\begin{figure}[!t]
	\begin{center}
		\includegraphics[width=0.65\textwidth]{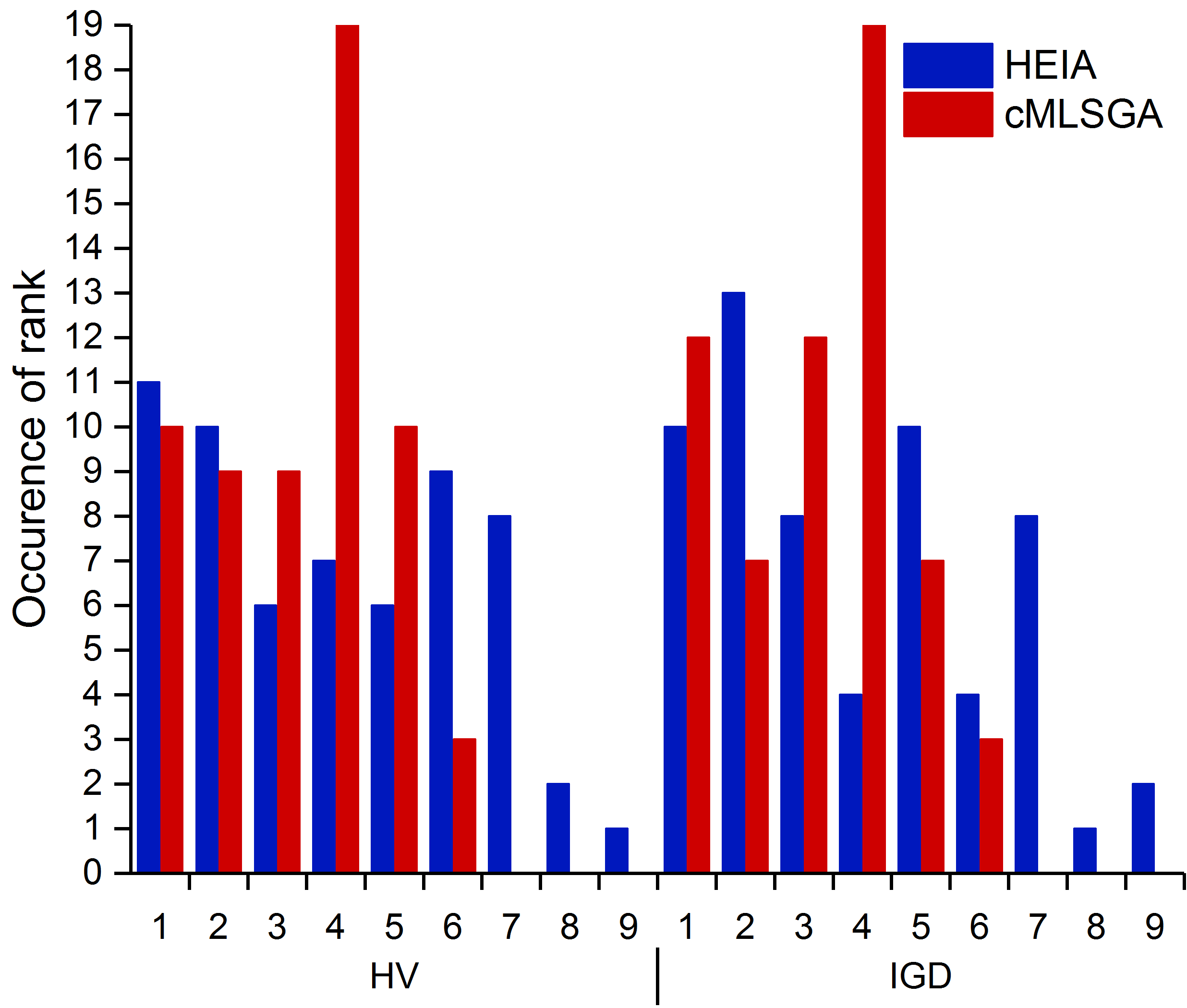}
		\caption{Occurrence of ranks for HEIA and cMLSGA algorithms on two-objective problems.}
		\label{fig:Rank_Occ}
		
	\end{center}
\end{figure}

\begin{figure}[!t]
	\begin{center}
		\includegraphics[width=0.65\textwidth]{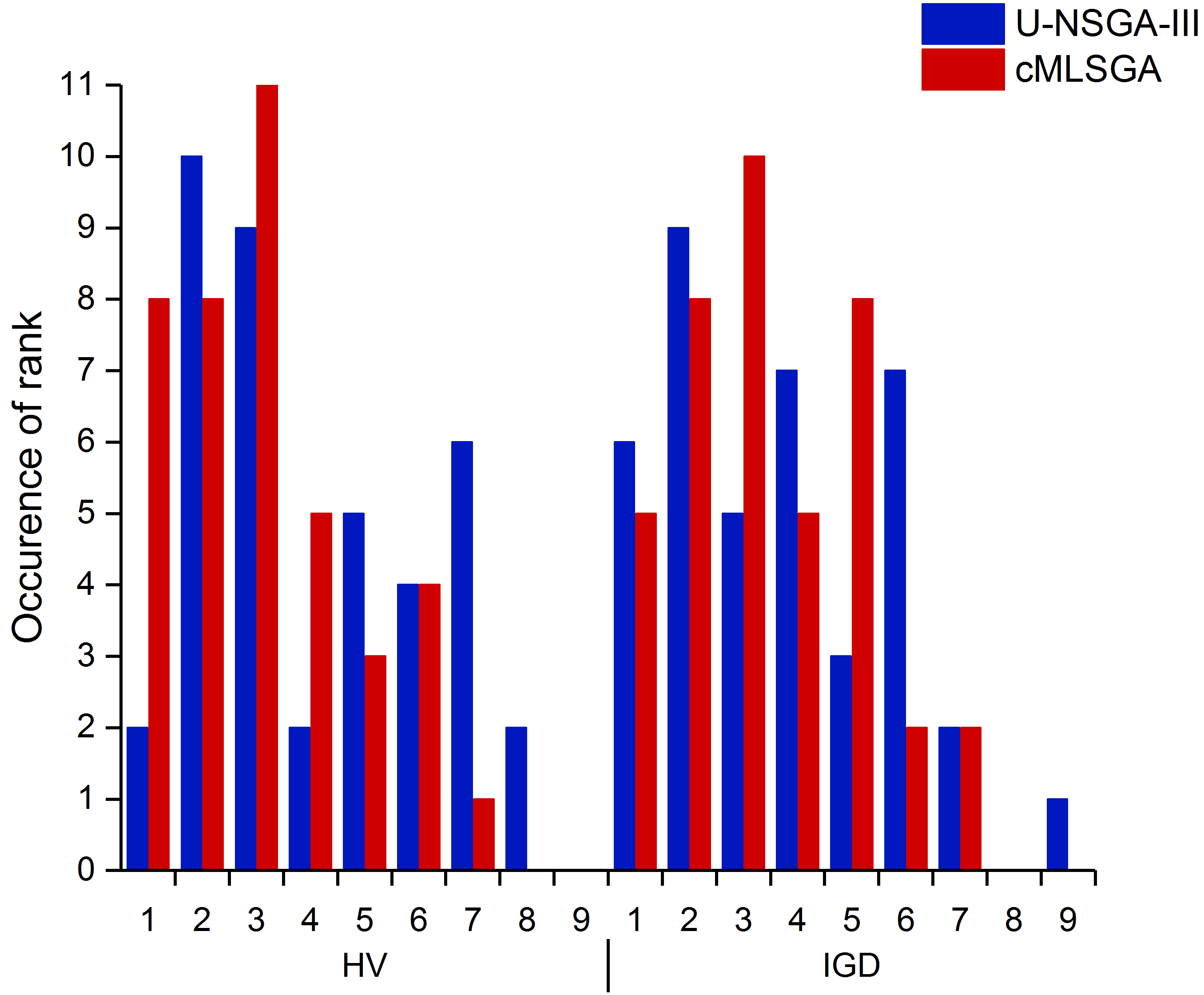}
		\caption{Occurrence of ranks for U-NSGA-III and cMLSGA algorithms on three objective problems.}
		\label{fig:Rank_Occ_3}
		
	\end{center}
\end{figure}

\begin{table}[!t]
	\caption{The rankings for the 9 genetic algorithms according to the average performance on different two-objective problem categories for IGD/HV indicator}
	\label{Tab:Rank}
	\begin{center}
		\begin{scriptsize}
			\begin{tabular}{>{\centering}m{0.8cm}|>{\centering}m{1cm}|>{\centering}m{1cm}|>{\centering}m{1cm}|>{\centering}m{1cm}|>{\centering}m{0.5cm}|>{\centering}m{0.5cm}|>{\centering}m{0.5cm}|>{\centering}m{0.6cm}|>{\centering}m{0.9cm}}			
				\hline 
				\textbf{Cat.} & \textbf{UNSGA-III} & \textbf{MOEA/D}     & \textbf{MOEA/D PSF} & \textbf{MOEA/D MSF}  & \textbf{IBEA} & \textbf{MTS} & \textbf{BCE}       & \textbf{HEIA}        & \textbf{cMLSGA}      \tabularnewline \hline \hline 
				\textbf{I}        & 4.60/ 2.60       & \cellcolor{redbox}3.00/ 3.80          & 4.60/ 5.00         & 6.60/ 7.20          & 8.40/ 6.80     & 7.00/ 8.60    & 7.40/ 7.40          & \cellcolor{yellowbox}2.40/ 2.60            & \cellcolor{greenbox}1.00/ 1.00   \tabularnewline \hline
				\textbf{II}       & 7.00/ 7.33       & \cellcolor{redbox}3.00/ 3.33          & 4.67/ 4.00         & 3.33/ 3.67          & 8.00/ 9.00     & 9.00/ 7.33    & \cellcolor{yellowbox} 2.67/ 3.67          & \cellcolor{greenbox}1.33/ 1.00   & 6.00/ 5.67            \tabularnewline \hline
				\textbf{III}      & \cellcolor{redbox} 3.14/ 3.29       & 4.86/ 5.00          & 7.57/ 7.29         & 5.43/ 5.29          & 6.43/ 6.71     & 6.86/ 6.14    & 5.57/ 7.00          & \cellcolor{greenbox}2.57/ 2.14   &\cellcolor{greenbox} 2.57/ 2.14   \tabularnewline \hline
				\textbf{IV}       & \cellcolor{yellowbox}3.33/ 3.33       & \cellcolor{yellowbox}3.33/ 3.33          & 5.67/ 5.67         & 5.33/ 5.00          & 8.00/ 6.33     & 6.33/ 7.67    & 7.67/ 8.67          & \cellcolor{greenbox}1.00/ 1.00   & 4.33/ 4.00            \tabularnewline \hline
				\textbf{V}        & 4.25/ 5.25       & 5.50/ 4.50          & 5.25/ 5.25         & 5.00/ 4.25          & 7.75/ 6.75     & 7.00/ 7.75    & \cellcolor{yellowbox}3.25/ 3.75 & \cellcolor{redbox}3.75/ 4.50            &\cellcolor{greenbox} 3.25/ 3.00   \tabularnewline \hline
				\textbf{VI}       & 5.30/ 4.50       & 5.00/ 5.00          & \cellcolor{yellowbox}2.90/ 2.90         & \cellcolor{greenbox}2.30/ 2.50 & 8.00/ 7.50     & 7.60/ 8.10    & 4.70/ 4.80          & 5.00/ 5.30            & \cellcolor{redbox}4.20/ 4.40            \tabularnewline \hline
				\textbf{VII}      & 6.00/ 3.33       & 5.67/ 6.33          & 6.33/ 5.67         & 5.00/ 6.00          & 6.33/ 8.00     & 8.00/ 6.33    & \cellcolor{yellowbox}4.00/ 2.67          & \cellcolor{redbox}2.33/ 4.67            &\cellcolor{greenbox} 1.33/ 2.00   \tabularnewline \hline
				\textbf{VIII}     & 4.25/ 3.75       & 6.75/ 6.00          & 7.25/ 7.00         & 6.00/ 6.75          & 6.25/ 8.75     & 8.50/ 6.50    & \cellcolor{greenbox}1.00/ 1.00 &\cellcolor{redbox} 2.50/ 3.25            & \cellcolor{yellowbox} 2.50/ 2.00            \tabularnewline \hline
				\textbf{IX}       & 7.33/ 7.00       &\cellcolor{greenbox}1.00/ 1.33 & \cellcolor{redbox}3.67/ 3.00         &\cellcolor{yellowbox} 2.67/ 2.67          & 6.00/ 9.00     & 9.00/ 6.00    & 7.67/ 8.00          & 4.00/ 4.67            & 3.67/ 3.33            
				\tabularnewline \hline
				\textbf{X}    &  3.83/ 4.17&	5.00/ 4.83&	5.00/ 5.42&	\cellcolor{redbox}3.83/ 3.83&	8.83/ 8.75&	6.17/ 5.67&	\cellcolor{greenbox}3.33/ 3.50&	5.67/ 5.33&	\cellcolor{greenbox}3.33/ 3.50					 
				\tabularnewline \hline
				\textbf{XI}    &  \cellcolor{redbox} 4.50/ 4.17	&5.75/ 3.75&	5.75/ 5.75&	5.08/ 5.25&	7.58/ 8.08&	6.67/ 6.50&	\cellcolor{yellowbox}2.50/ 4.00&	5.00/ 3.83&\cellcolor{greenbox}	2.17/ 3.67
				\tabularnewline \hline
				\textbf{XII}     &  \cellcolor{redbox}4.00/ 4.00&\cellcolor{greenbox}	4.50/ 3.33&	4.83/ 5.58	&\cellcolor{greenbox}4.17/ 3.67&	8.67/ 8.42	&4.17/ 4.00&	4.50/ 4.67&	6.17/ 6.83&	4.00/ 4.50					 					
				\tabularnewline \hline \hline
				\textbf{Overall}  & \cellcolor{redbox}4.617/ 4.233    & \cellcolor{redbox}4.642/ 4.342        & 5.158/ 5.158       & 4.425/ 4.508        & 7.642/ 7.625   & 7.000/ 6.867  & 4.467/ 4.917        & \cellcolor{yellowbox}3.867/ 4.033 & \cellcolor{greenbox}3.183/ 3.317 \tabularnewline \hline
				\textbf{Std}      & 2.244/ 2.348	&2.564/ 2.353&	2.294/ 2.369	&2.278/ 2.469&	1.435/ 1.488	&2.106/ 2.164	&2.735/ 2.691	&2.284/ 2.295	&1.455/ 1.466					
				\tabularnewline \hline \hline
				\multicolumn{10}{>{\centering}m{11.5cm}}{\textit{
						The best algorithm in each category is highlighted in green, the second and the third best are highlighted in orange and red respectively.}}
			\end{tabular}
		\end{scriptsize}
	\end{center}
\end{table}

\begin{table}[!t]
	\caption{The rankings for the 9 genetic algorithms according to the average performance on different three-objective problem categories for IGD/HV indicator}
	\label{Tab:Rank_3}
	\begin{center}
		\begin{scriptsize}
			\begin{tabular}{>{\centering}m{0.8cm}|>{\centering}m{1cm}|>{\centering}m{1cm}|>{\centering}m{1cm}|>{\centering}m{1cm}|>{\centering}m{0.5cm}|>{\centering}m{0.5cm}|>{\centering}m{0.5cm}|>{\centering}m{0.6cm}|>{\centering}m{0.9cm}}			
				\hline 
				\textbf{Cat.} & \textbf{UNSGA-III} & \textbf{MOEA/D}     & \textbf{MOEA/D PSF} & \textbf{MOEA/D MSF}  & \textbf{IBEA} & \textbf{MTS} & \textbf{BCE}       & \textbf{HEIA}        & \textbf{cMLSGA}      \tabularnewline \hline \hline 
				\textbf{I}       & \cellcolor{redbox}3.31/ 3.85   & \cellcolor{yellowbox}3.77/ 3.15   & 4.92/ 5.77                         & 6.23/ 6.23                         & 6.92/ 7.38    & 5.77/ 4.31     & 7.15/ 7.69                         & 4.00/ 3.46                         & \cellcolor{greenbox}2.92/ 3.15   \tabularnewline  \hline
				\textbf{IV}      & \cellcolor{yellowbox}2.67/ 2.33   & \cellcolor{redbox}3.67/ 4.67   & 4.67/ 4.33                         & 5.67/ 6.67                         & 9.00/ 8.67    & 4.00/ 4.67     & 7.33/ 8.33                         & \cellcolor{greenbox}3.33/ 1.33 & 4.67/ 4.00                           \tabularnewline  \hline
				\textbf{V}     & \cellcolor{greenbox}2.67/ 2.00   & 7.00/ 4.67                           & 5.00/ 4.67                         & 6.33/ 5.33                         & 4.00/ 6.67    & 7.00/ 7.33     & 5.00/ 8.00                         & \cellcolor{redbox}4.33/ 4.00 & \cellcolor{yellowbox}3.67/ 2.33   \tabularnewline  \hline
				\textbf{VI}      & 5.83/ 6.50                           & \cellcolor{redbox}4.00/ 4.50   & \cellcolor{greenbox}1.17/ 1.33 & \cellcolor{yellowbox}2.17/ 2.17 & 6.67/ 6.00    & 8.83/ 8.83     & 6.33/ 6.67                         & 5.00/ 4.67                         & 5.00/ 4.33                           \tabularnewline  \hline
				\textbf{VII}       & \cellcolor{redbox}4.40/ 4.00   & 5.20/ 5.10                           & 5.20/ 5.50                         & 4.40/ 5.20                         & 5.60/ 6.30    & 6.40/ 6.40     & \cellcolor{greenbox}3.40/ 3.30 & 6.80/ 5.80                         & \cellcolor{yellowbox}3.60/ 3.40   \tabularnewline  \hline
				\textbf{IX}      & 5.00/ 5.00                           & \cellcolor{yellowbox}1.00/ 3.00   & \cellcolor{greenbox}2.00/ 1.00 & \cellcolor{redbox}4.00/ 2.00 & 6.00/ 7.00    & 8.00/ 8.00     & 9.00/ 9.00                         & 3.00/ 4.00                         & 7.00/ 6.00                           
				\tabularnewline  \hline
				\textbf{X}      & 4.67/ 5.67&	8.00/ 6.00&	5.33/ 6.83&	5.00/ 5.67&	9.00/ 8.50&	\cellcolor{yellowbox}3.67/ 3.67&	\cellcolor{yellowbox}3.00/ 4.00&	4.67/ 3.00&	\cellcolor{greenbox}1.67/ 1.67										
				\tabularnewline  \hline
				\textbf{XI}      &\cellcolor{yellowbox} 1.00/ 3.33&	5.67/ 5.33	&7.33/ 7.17&	7.33/ 7.17&	7.33/ 7.17&	4.83/ 4.67&	6.50/ 6.50&	\cellcolor{redbox}3.00/ 2.33&	\cellcolor{greenbox}2.00/ 1.33										   
				\tabularnewline  \hline
				\textbf{XII}      &\cellcolor{redbox}3.00/ 3.33&	8.33/ 7.17&	5.00/ 6.33	&4.67/ 6.17&	8.33/ 7.67&	4.67/ 4.67&	5.00/ 5.33	&\cellcolor{yellowbox}3.33/ 2.67& \cellcolor{greenbox}	2.67/ 1.67										       
				\tabularnewline  \hline \hline
				\textbf{Overall} & \cellcolor{yellowbox}3.675/ 4.100	&4.950/ 4.500&	4.525/ 4.988	&5.175/ 5.375&	6.925/ 7.163&	6.013/ 5.600	&5.938/ 6.550&	\cellcolor{redbox}4.375/ 3.650&	\cellcolor{greenbox}3.425/ 3.075					
				\tabularnewline  \hline
				\textbf{std}     & 2.030/ 2.083&	2.629/ 2.188	&2.786/ 2.628&	2.293/ 2.296&	2.020/ 1.690	&2.481/ 2.615	&2.760/ 2.405	&2.070/ 2.264&	1.641/ 1.649					
				\tabularnewline  \hline \hline
				
				\multicolumn{10}{>{\centering}m{11.5cm}}{\textit{
						The best algorithm in each category is highlighted in green, the second and the third best are highlighted in orange and red respectively.}}
			\end{tabular}
		\end{scriptsize}
	\end{center}
\end{table}

The performance of the best developed cMLSGA variant is compared to 8 selected current state-of-the-art algorithms in order to evaluate the potential of the cMLSGA approach as a general solver. In the case of two-objectives, the cMLSGA\footnotesize \_HEIA\_MOEA/D-MSF \normalsize variant is utilised, whereas for more objectives the cMLSGA\footnotesize \_MOEA/D-MSF\_U-NSGA-III \normalsize is compared, due to the outstanding performance of U-NSGA-III on three-objective problems, and the lower effectiveness of HEIA on them. This is presented in the form of an average ranking on each category of problem in Table \ref{Tab:Rank} for two-objective problems and \ref{Tab:Rank_3} for three-objective problems. From the two-objective ranking it can be observed that the best solver on average is cMLSGA, 3.183/3.317 for the IGD/HV metrics, while HEIA comes second with 3.867/4.033 for the IGD/HV indicators. Higher differences in the HV scores indicates that cMLSGA is more likely to promote high diversity than HEIA. In addition cMLSGA is shown to be more general than HEIA, which has a higher deviation in performance with larger standard deviations of its position in the rankings presented in Table \ref{Tab:Rank}, 2.284 for IGD and 2.295 for HV, compared to cMLSGA, 1.455 for IGD and 1.466 for HV. This is illustrated in Fig. \ref{fig:Rank_Occ} for the two objective problems, where the occurrence of each rank is presented for the HEIA and cMLSGA algorithms. HEIA and cMLSGA are often the top performers, rank 1, but HEIA is also regularly the second-best algorithm on IGD, 13 times compared to 7 times for cMLSGA. However, when HEIA shows low performance on a problem it performs very poorly with rank 7 eight times for both indicators; rank 8 one time for IGD and two times for HV and rank 9 two times for IGD and one time for HV. Whereas the lowest rank for cMLSGA is 6 and it has a high occurrence in the 3$^{rd}$, 4$^{th}$ and 5$^{th}$ positions; demonstrating that cMLSGA is the best general solver. 

Similar behaviour can be observed for the three-objective problems. From the corresponding ranking, Table \ref{Tab:Rank_3}, it can be observed that the best solver on average is cMLSGA and U-NSGA-III comes second. The IGD rankings are 3.425 for cMLSGA and 3.675 for U-NSGA-III, and 3.075 and 4.100 according to HV. Similarly, to the bi-objective cases, cMLSGA also shows lower standard deviations than the second best-on-average algorithm, U-NSGA-III, further proving its high generality. Comparing the occurrence of each rank for U-NSGA-III and cMLSGA, illustrated on Fig. \ref{fig:Rank_Occ_3}, it can be seen that cMLSGA is again more likely to avoid poor performance. In this case, both algorithms are the top performers an approximately equal number of times for IGD, but cMLSGA is significantly more likely to exhibit the highest performance according to the HV metric. Furthermore, U-NSGA-III is more likely to be second, 9 and 10 times with IGD and HV respectively compared to 8 times for cMLSGA with both indicators. In contrast, cMLSGA is significantly less likely to perform poorly, as in the worst case scenario it is 7$^{th}$ for both indicators, whereas for U-NSGA-III the lowest position is 9$^{th}$ for IGD and 8$^{th}$ for HV. Furthermore, U-NSGA-III is more likely than cMLSGA to achieve ranks 6 and 7. Interestingly, both algorithms are unlikely to be 1$^{st}$ or last, which occurs less than two times for cMLSGA and U-NSGA-III, and are showing a distribution reflecting a general performance. This indicates a high generality and universality of both solutions which are often outperformed by specialist-solvers on their preferred cases. 

cMLSGA is bounded by the performance of the implemented algorithms and therefore their selection remains important. A self-adaptive variant may be considered in the future but currently the HEIA\_MEOA/D-MSF combination is suggested for two-objective problems and MOEA/D-MSF\_U-NSGA-III for more objectives, as these methodologies are the most distinct and show the most “general” behaviour on corresponding cases. By combining the advantages of the two methodologies selected at the individual level and utilising diversity-first approaches, derived from the multi-level selection mechanisms, the chance of success of cMLSGA is still significantly higher than other algorithms, especially in cases where there is no knowledge about the optimisation space. The profile is more similar to that of U-NSGA-III than the rest of the state-of-the-art, as this algorithm also shows a strong general performance and especially high performance on the HV diversity metric. This combination provides confidence that cMLSGA will provide good results on cases where there is no a priori knowledge of the problem characteristic or where there are multiple dominating characteristics; the case for many real-world problems. In cases where there is a priori knowledge of the problem the provided supporting data, showing the results for all of the variants of cMLSGA, will allow determination of the strongest variant for a given problem type.	

\section{Conclusions}
Current real-world problems have been shown to be more complex than the available evolutionary computation benchmarking problems, requiring algorithms tested on constrained and imbalanced problems that require high diversity. In addition, when solving practical optimisation problems, there is often no a priori knowledge about the problem set, meaning that specialist algorithms can’t be selected in advance. To develop an algorithm with both high diversity and general properties, this paper presents a novel co-evolutionary approach. The novel approach is the first to develop an approach dependant on competition between collectives and the additional mechanisms are added to the Multi-Level Selection Genetic Algorithm, designated cMLSGA. A lower communication rate between sub-populations than in other  co-evolutionary approaches and a split in the fitness definition between individual and collective levels allows an independence of search between the different collectives. The new methodology promotes a diversity first, convergence second behaviour seen in previous iterations of this algorithm, improving performance on highly discontinuous, constrained and irregular problems. An investigation into the optimum pairing of algorithms in the co-evolutionary approach provides general guidance for developers of algorithms utilising these approaches, showing that the diversity of the co-evolutionary mechanisms is more important than their individual performance. Based on this pairing the resulting cMLSGA is shown to be the best currently existing general-solver Evolutionary Algorithm. This greatly increases cMLSGA’s applicability to real-world problems, especially on cases where the problem characteristics are not known or where they are dominated by more than one characteristic.

\section*{Acknowledgements}
This work was sponsored by Lloyd's Register Foundation.

	\bibliographystyle{elsarticle-num}
	\bibliography{4_cMLSGA_Bibliography}

\end{document}